\documentclass{article}
\usepackage{spconf,amsmath,amssymb,amsfonts}
\usepackage{algorithm, algorithmic}
\usepackage{comment}
\usepackage{etoolbox}
\usepackage{booktabs}
\usepackage{adjustbox}
\usepackage{url}
\makeatletter
\patchcmd{\@makecaption}
  {\scshape}
  {}
  {}
  {}
\makeatother
\usepackage{graphicx}
\usepackage{textcomp}
\usepackage{xcolor}
\usepackage{multirow}
\usepackage[skip=2pt]{caption}
\def\BibTeX{{\rm B\kern-.05em{\sc i\kern-.025em b}\kern-.08em
    T\kern-.1667em\lower.7ex\hbox{E}\kern-.125emX}}

\title{Privacy-Preserving Federated Action Recognition via Differentially Private Selective Tuning and Efficient Communication}

\name{
  \begin{tabular}{c}
    Idris Zakariyya$^{1}$, Pai Chet Ng$^{2}$, Kaushik Bhargav Sivangi$^{1}$, \\
    S. Mohammad Sheikholeslami$^{3}$, Konstantinos N. Plataniotis$^{3}$, Fani Deligianni$^{1}$
  \end{tabular}
}

\address{
  \begin{tabular}{c}
    $^{1}$ School of Computing Science, University of Glasgow, G12 8QQ, Glasgow, UK \\
    $^{2}$ Infocomm Technology Cluster, Singapore Institute of Technology, Singapore \\
    $^{3}$ The Edward S. Rogers Sr. Dept. of Electrical and Computer Engineering, University of Toronto, Canada \\
    Emails: \{idris.zakariyya, fani.deligianni\}@glasgow.ac.uk, paichet.ng@singaporetech.edu.sg
  \end{tabular}
}

\begin{document}
%\ninept
%
\maketitle
\begin{abstract}
Federated video action recognition enables collaborative model training without sharing raw video data, yet remains vulnerable to two key challenges: \textit{model exposure} and \textit{communication overhead}. Gradients exchanged between clients and the server can leak private motion patterns, while full-model synchronization of high-dimensional video networks causes significant bandwidth and communication costs. To address these issues, we propose \textit{Federated Differential Privacy with Selective Tuning and Efficient Communication for Action Recognition}, namely \textit{FedDP-STECAR}. Our \textit{FedDP-STECAR} framework selectively fine-tunes and perturbs only a small subset of task-relevant layers under Differential Privacy (DP), reducing the surface of information leakage while preserving temporal coherence in video features. By transmitting only the tuned layers during aggregation, communication traffic is reduced by over 99\% compared to full-model updates. Experiments on the UCF-101 dataset using the MViT-B-16x4 transformer show that \textit{FedDP-STECAR} achieves up to \textbf{70.2\% higher accuracy} under strict privacy ($\epsilon=0.65$) in centralized settings and \textbf{48\% faster training} with \textbf{73.1\% accuracy} in federated setups, enabling scalable and privacy-preserving video action recognition. Code available at https://github.com/izakariyya/mvit-federated-videodp

\end{abstract}
\section{Introduction}
\label{sec:intro}

%The global population is projected to surpass 9.5 billion by 2050, with a rapidly growing elderly demographic~\cite{world_pop}. This trend highlights the increasing need for intelligent healthcare and assistive technologies that can understand and monitor human actions in daily life. 

\begin{figure}
    \centering
    \includegraphics[width=.95\columnwidth]{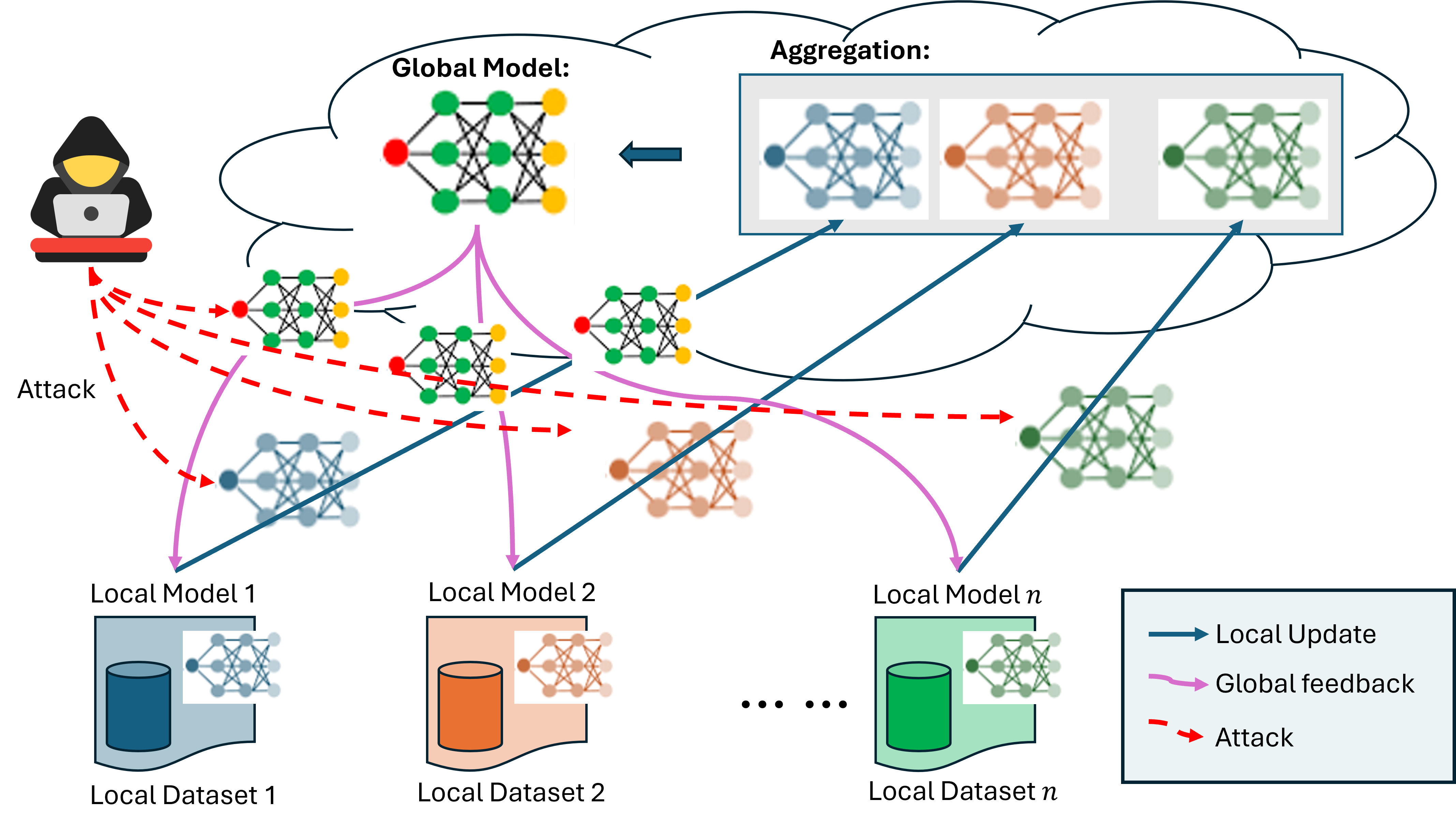}
    \caption{\textbf{Vulnerabilities in federated video learning.}  
    (a) \textit{Model exposure:} Shared gradients may reveal sensitive motion motion or identity.  
    (b) \textit{Communication overhead:} Synchronizing large video models cause heavy traffic and latency.}
    \label{fig:problem}
\end{figure}

Video-based \textit{action recognition} using deep neural networks (DNNs) has become an essential \cite{du2025skeleton}\cite{boitel2025mist}, as it enables accurate and contextual interpretation of user behaviors in real-world environments.
While centralized DNNs achieve remarkable accuracy in action recognition, their reliance on large-scale video data raises serious privacy concerns. Federated Learning (FL) enables collaborative model training without centralizing data, but, as illustrated in Fig.~\ref{fig:problem}, FL remains vulnerable to two fundamental limitations: (\textit{i}) \textit{model exposure}, where shared gradients can be exploited to reconstruct visual content or infer user actions, and (\textit{ii}) \textit{communication overhead}, where synchronizing full video models across clients saturates bandwidth and increases energy cost. These challenges hinder the secure and efficient deployment of FL-based action recognition in real-world healthcare and surveillance systems.
Differential Privacy (DP) \cite{dwork2014algorithmic} offers a principled way to limit information leakage by adding calibrated noise to the learning process. However, integrating DP into federated video learning is non-trivial: strong privacy budgets ($\varepsilon \!\downarrow$) often degrade model accuracy, uniform noise across all layers inflates communication cost, and temporal correlations in videos amplify privacy sensitivity. Hence, achieving an optimal balance among privacy, accuracy, and communication remains an open challenge.

Existing works only partially address these issues. 
DP-FedAvg~\cite{geyer2017differentially} and FedCDP~\cite{agarwal2021federated} introduce noise-based DP aggregation but overlook communication efficiency and layer-wise optimization. 
In video tasks, Luo \textit{et al.}~\cite{10484452} propose multi-clip DP-SGD for video-level privacy, Schneider \textit{et al.}~\cite{schneider2024activity} apply Masked DP to anonymized spatio-temporal regions, and Nken \textit{et al.}~\cite{nken2025video} use DP projections for visual privacy. 
While effective in centralized scenarios, these methods are difficult to scale in non-IID federated environments and often require heavy noise injection, leading to unstable convergence and degraded temporal fidelity. 
To date, no framework jointly optimizes privacy, accuracy, and communication for federated video action recognition.

To overcome these limitations, we propose \textit{Federated Differential Privacy with Selective Tuning and Efficient Communication for Action Recognition (FedDP-STECAR)}.  
Our \textit{FedDP-STECAR} mitigates the two key vulnerabilities in federated video learning:
(\textit{i}) \textit{Model exposure} is reduced by selectively fine-tuning and perturbing only high-impact layers, limiting information leakage while preserving temporal discriminability; and
(\textit{ii}) \textit{Communication overhead} is minimized by transmitting only the tuned layers per round, reducing bandwidth and communication cost.
% This aggregation-agnostic design enables secure, scalable, and resource-efficient training under strict privacy budgets, and generalizes across transformer-based video backbones.
Our contributions are summarized as follows:
\begin{enumerate}
     \item Our selective fine-tuning strategy confines DP noise to task-relevant parameters, reducing unnecessary perturbations and preserving temporal coherence in video features. Under a strict privacy budget ($\epsilon = 0.65$), the selective DP model achieves up to {70.2\% higher accuracy} than full fine-tuning baselines in centralized training.

    \item In federated settings, \textit{FedDP-STECAR} reduces total {runtime by 48\%} and {communication traffic by 99\%} while maintaining high recognition accuracy (73.1\%) at $\epsilon = 1.33$, demonstrating its scalability and practicality for privacy-preserving video action recognition.

    \item The framework further integrates \textit{Top-Level Sampling (TLS)} for privacy amplification via sampling without replacement. While TLS improves model utility under strict privacy in centralized setups, it introduces a minor trade-off in federated settings, achieving comparable accuracy with only slight performance degradation.
\end{enumerate}

\section{Preliminaries and Related Work} \label{sec:background}
%Privacy-preserving learning is critical for video-based action recognition, where raw data may reveal sensitive personal information.

FL enables decentralized training without sharing raw data, yet exchanged gradients can still leak private details.
DP mitigates this risk by injecting calibrated noise into gradient updates, but integrating DP with FL introduces two key challenges:
(i) utility degradation from excessive noise,
(ii) high communication overhead from frequent model exchanges.

Formally, given $K$ clients each with a private dataset 
$\mathcal{D}_k = \{(x_i^{(k)}, y_i^{(k)})\}$,
the goal is to minimize the expected empirical risk under privacy and communication constraints:
\begin{equation}
\label{eq:objective}
\min_{\theta}~ \mathbb{E}_{k}\big[F_k(\theta)\big]
\quad \text{s.t.} \quad (\varepsilon, \delta)\text{-DP and limited bandwidth.}
\end{equation}
This work addresses the question:
\textit{How can federated video models achieve strong DP protection while maintaining accuracy and minimizing communication cost?}
To answer this, we revisit the principles of DP and FL, analyze existing methods, and introduce our proposed \textit{FedDP-STECAR} framework.

\subsection{Differential Privacy: Formulation and Related Work}
\noindent
\textbf{Differential Privacy. }
The DP mechanism is a statistical framework that safeguards sensitive data during DNN model training by injecting carefully calibrated noise into computational processes ~\cite{dwork2014algorithmic}. Formally, a randomized mechanism $\mathcal{G}$ satisfies $(\varepsilon, \delta)$-DP if,
for any two adjacent datasets $X_1$ and $X_2$ differing by at most one record,
and for all measurable subsets $\mathcal{T}$ of the output space,
\begin{equation}
\label{eq:dp_def}
\mathbb{P}[\mathcal{G}(X_1) \in \mathcal{T}] 
\le e^{\varepsilon}\,\mathbb{P}[\mathcal{G}(X_2) \in \mathcal{T}] + \delta,
\end{equation}
where $(\varepsilon, \delta)$ bound the privacy loss.

In deep learning, the most widely used mechanism is \textbf{DP-SGD}~\cite{abadi2016deep},
which protects privacy at the gradient level.
For a batch $\mathcal{B}_t$ at iteration $t$, per-sample gradients are clipped
and Gaussian noise is added:
\begin{equation}
\label{eq:dp_sgd}
\hat{g}_t = 
\frac{1}{|\mathcal{B}_t|}\sum_{i\in\mathcal{B}_t}
\frac{g_i}{\max(1,\|g_i\|_2/C)} + \mathcal{N}(0,\sigma^2C^2 I),
\end{equation}
where $C$ is the clipping threshold and $\sigma$ the noise multiplier.
The model update then follows $\theta_{t+1} = \theta_t - \eta \hat{g}_t$.

Sampling without replacement provides additional \textit{privacy amplification}~\cite{balle2018privacy}.
For a sampling ratio $q = B/N$ and $E$ local epochs,
the approximate privacy cost can be expressed as:
\begin{equation}
\label{eq:eps_privacy}
\varepsilon_{\text{priv}}(E,q,\sigma,\delta)
= \frac{q}{\sigma}\sqrt{2E\ln(1/\delta)}.
\end{equation}

\noindent 
\textbf{Related Work on Differential Privacy. }
Recent studies have explored enhancing DP-SGD through
adaptive clipping~\cite{andrew2021differentially}, efficient gradient accounting~\cite{dupuy2022efficient},
and dynamic noise scheduling~\cite{boenisch2024have}.
While these methods improve privacy-utility trade-offs in image recognition or NLP,
they remain focused on centralized settings.
In video applications, Luo \textit{et al.}~\cite{10484452} introduced a
multi-clip DP-SGD scheme for video-level protection,
and Nken \textit{et al.}~\cite{nken2025video} proposed projection-based privacy for human motion data.
However, these methods require high privacy budgets ($\varepsilon>5$)
and ignore distributed training and communication efficiency.

\subsection{Federated Learning: Formulation and Related Work}
FL~\cite{mcmahan2017communication}
enables model training across distributed clients without centralizing data.
The global objective is defined as:
\begin{equation}
\label{eq:fl_obj}
\min_{\theta} F(\theta) 
= \sum_{k=1}^{K} p_k F_k(\theta), 
\quad p_k = \frac{n_k}{\sum_j n_j},
\end{equation}
where $F_k(\theta)$ is the local loss at client $k$.
Each client performs local optimization:
\begin{equation}
\label{eq:local_update}
\theta_{k,t+1} = \theta_t - \eta\nabla F_k(\theta_t),
\end{equation}
and transmits model updates $\Delta_k^t = \theta_{k,t+1}-\theta_t$ to the server.
The server aggregates updates from participating clients $S_t$:
\begin{equation}
\label{eq:global_update}
\theta_{t+1} = \theta_t + \sum_{k\in S_t} p_k\,\mathcal{A}(\Delta_k^t),
\end{equation}
where $\mathcal{A}(\cdot)$ denotes the aggregation operator.
For instance, $\mathcal{A}(\Delta_k^t)=\Delta_k^t$ for
{FedAvg}~\cite{mcmahan2017communication} and
$\mathcal{A}(\Delta_k^t)=\Delta_k^t/\tau_k$ for
{FedNova}~\cite{wang2020tackling}, where $\tau_k$ is the local epoch count.

\noindent
\textbf{Related Work on Federated Learning. }
Numerous extensions of FL have been proposed to improve
heterogeneity handling, convergence, and personalization.
FedProx~\cite{li2020federated} mitigates client drift with a proximal term,
FedDyn~\cite{acar2021federated} stabilizes updates using dynamic regularization,
and FedPer~\cite{arivazhagan2019federated} introduces layer-wise personalization.
While these methods enhance convergence or fairness, 
they generally do not provide formal privacy guarantees.
For video-based tasks, the high-dimensional feature representations
and temporal redundancy exacerbate both communication and privacy costs.

\subsection{Integration of FL and DP}
Several works have explored combining FL and DP to achieve privacy-preserving distributed learning.
DP-FedAvg~\cite{geyer2017differentially} introduced per-round Gaussian noise at the server level,
while FedCDP~\cite{agarwal2021federated} applied client-level composition for tighter accounting.
Hybrid cryptographic-DP approaches~\cite{fouzar2023novel}
further enhance security but incur heavy computational costs.
Despite these advances, existing DP-FL methods often:
(1) inject uniform noise across all layers,  
(2) overlook layer sensitivity and sampling strategies, and  
(3) do not explicitly balance privacy, accuracy, and communication efficiency.
Thus, their utility drops sharply under stringent privacy budgets.

%These limitations motivate a new design that integrates selective parameter tuning, privacy-amplified sampling, and communication-aware regulation within a unified framework.
%To address the limitations, we propose \textit{FedDP-STECAR} that achieves privacy-preserving and communication-efficient federated learning under strong DP constraints.
%By jointly optimizing privacy, utility, and communication efficiency,
%\textit{FedDP-STECAR} advances differentially private video recognition
%toward practical deployment in federated environments.

\section{Proposed Framework: FedDP-STECAR} \label{sec:proposed}
This section introduces the proposed \textit{FedDP-STECAR} framework, which achieves privacy-preserving and communication-efficient federated learning through \textit{Differentially Private Selective Fine-tuning}. 
On each client, only a subset of task-relevant parameters $\theta_t \subset \Theta$ is updated using a DP-SGD optimizer with TLS, confining noise to critical layers and reducing computation. The server then aggregates the masked client updates $\Delta_k^t$ in an \textit{aggregation-agnostic} manner, minimizing communication cost while preserving model utility.
% As illustrated in Fig.~\ref{fig:framework}, 

% \begin{figure}
%     \centering
%     \includegraphics[width=0.5\linewidth]{...}
%     \caption{Ours FedDP-STECAR: To be added}
%     \label{fig:feddp}
% \end{figure}
% \subsection{DP with Selective Fine-tuning and Top-Level Sampling}

\noindent
\textbf{Selective Tuning. }
\textit{FedDP-STECAR} partitions model parameters into frozen and trainable subsets:
\begin{equation}
\Theta = [\theta_f, \theta_t], \quad \text{where } \nabla\theta_f = 0
\end{equation}
where $\theta_t \subset \Theta$ denotes trainable parameters and $\theta_f$ remain fixed.

DP-SGD operates only on $\theta_t$, restricting sensitivity to a smaller subspace and limiting noise propagation.
Given a mini-batch $\mathcal{B}_t$, the clipped and noise-perturbed gradient on client $k$ is
\begin{equation}
\label{eq:dp_grad}
\hat{g}^{(k)}_t = 
\frac{1}{|\mathcal{B}_t|} \sum_{i\in\mathcal{B}_t} 
\frac{g_{i,t}^{(k)}}{\max(1,\|g_{i,t}^{(k)}\|_2/C)} + 
\mathcal{N}(0,\sigma^2C^2\mathbf{I}),
\end{equation}
where $C$ is the clipping bound and $\sigma$ the noise multiplier.
The update is applied only to $\theta_t$:
\begin{equation}
\theta_t \leftarrow \theta_t - \eta\,\hat{g}^{(k)}_t.
\end{equation}
thereby confining DP noise to critical layers and improving both privacy and efficiency.

\medskip
\noindent
\textbf{Top-Level Sampling (TLS).}
TLS ensures each sample is used once per local epoch through sampling without replacement, reducing repeated gradient exposure and strengthening privacy.
For sampling ratio $q=B/N$ and local epochs $E$, the cumulative DP cost follows:
\begin{equation}
\label{eq:eps_eff_method}
\varepsilon_{\text{TLS}} =
\frac{q}{\sigma}\sqrt{2E\ln(1/\delta)}.
\end{equation}
TLS also enhances learning stability by maintaining temporal diversity within batches.

\subsection{Communication-efficient FL Update}
\noindent
\textbf{Aggregation-Agnostic Update. }
\textit{FedDP-STECAR} remains agnostic to the aggregation algorithm, supporting 
both \textbf{FedAvg} and \textbf{FedNova}.
Each client $k$ computes a masked update $\Delta_k^t$ based on 
the trainable subset $\theta_t$:
\begin{equation}
\Delta_k^t = \mathrm{Mask}_{\theta_t}(w_{k,t+1} - w_t),
\end{equation}
where $\mathrm{Mask}_{\theta_t}(\cdot)$ retains only coordinates corresponding to $\theta_t$.  
The global aggregation at the server is written as:
\begin{equation}
\label{eq:agg}
w_{t+1} = w_t + \sum_{k\in S_t} p_k\,\mathcal{A}(\Delta_k^t),
\end{equation}
where $\mathcal{A}(\cdot)$ is the aggregation operator:
$\mathcal{A}(\Delta_k^t) = \Delta_k^t$ for FedAvg and 
$\mathcal{A}(\Delta_k^t) = \Delta_k^t/\tau_k$ for FedNova, 
and $p_k = n_k / \sum_{j\in S_t} n_j$.

\noindent
\textbf{Communication Efficiency. }
By transmitting only the selectively tuned layers $\theta_t$, 
the communication cost per client per round is significantly reduced:
\begin{equation}
C_{\text{comm}} = \frac{d_t}{d} B_f,
\end{equation}
where $B_f$ denotes the baseline bandwidth required to upload all model parameters. 
Combining this with the privacy term in~\eqref{eq:eps_eff_method} yields the 
joint objective to minimize privacy expenditure and communication overhead:
\begin{equation}
\label{eq:joint_obj}
\min_{w_t} \mathcal{L}(w_t) + \beta\,\varepsilon_{\text{eff}},
\end{equation}
where $\beta$ balances task accuracy and efficiency. 
The overall training process of FedDP-SsTEER is summarized in 
Algorithm~\ref{alg:feddp_stecar}. 
Each communication round consists of local TLS-based DP-SGD updates 
on the selected parameter subset and aggregation of masked updates at the server.

\begin{algorithm}[!b]
\caption{Our FedDP-STECAR}
\label{alg:feddp_stecar}
\begin{algorithmic}[1]
\REQUIRE 
$K$ clients with data $\{X_k\}$, total rounds $T$, local epochs $E$, 
batch size $B$, clipping bound $C$, noise multiplier $\sigma$, 
trainable subset $\theta_t$, and aggregation operator $\mathcal{A}(\cdot)$.
\ENSURE 
Global model $w_T$ after $T$ rounds.
\STATE Initialize $w_0$.
\FOR{each round $t = 0,1,\dots,T-1$}
    \STATE Server selects clients $S_t$ and broadcasts $w_t$.
    \FOR{each client $k \in S_t$ \textbf{in parallel}}
        \STATE Copy $w^{(k)} \!\gets\! w_t$, freeze $\theta_f$, train $\theta_t \!\subset\! \Theta$.
        \FOR{epoch $e=1$ \TO $E$}
            \STATE \textbf{TLS:} Shuffle $X_k$ to form mini-batches $\{\mathcal{B}_1,\dots,\mathcal{B}_M\}$.
            \FOR{each batch $\mathcal{B}_m$}
                \STATE Compute per-sample gradients $\{g_i\}_{i\in\mathcal{B}_m}$ on $\theta_t$.
                \STATE Clip and perturb gradients:
                \[
                \hat{g} \gets \frac{1}{|\mathcal{B}_m|}\sum_{i}\frac{g_i}{\max(1,\|g_i\|_2/C)}
                + \mathcal{N}(0,\sigma^2C^2I)
                \]
                \STATE Update $\theta_t \gets \theta_t - \eta\hat{g}$.
            \ENDFOR
        \ENDFOR
        \STATE Compute masked update $\Delta_k^t \!=\! \mathrm{Mask}_{\theta_t}(w^{(k)}\!-\!w_t)$ and send to server.
    \ENDFOR
    \STATE Server aggregates:
    \[
    w_{t+1} \!=\! w_t + \sum_{k\in S_t} p_k\,\mathcal{A}(\Delta_k^t)
    \]
\ENDFOR
\STATE \textbf{Return} global model $w_T$.
\end{algorithmic}
\end{algorithm}

\section{Dataset and Experimental Setup}
% This section outlines the dataset, experimental settings, and implementation details used to evaluate \textit{FedDP-STECAR}, focusing on its privacy, accuracy, and communication efficiency in video action recognition.

%The UCF-101 dataset ~\cite{soomro2012ucf101} is a comprehensive and widely adopted benchmark for action recognition in video analysis. It consists of 13,320 video clips, each annotated into one of 101 action classes, covering a broad spectrum of human activities. These classes encompass a diverse array of actions, ranging from daily activities such as \textit{ApplyEyeMakeup} and \textit{BrushingTeeth} to sports-related actions like \textit{PlayingTennis} and \textit{BasketballDunk}. The dataset is particularly notable for its realistic and challenging video content, which is sourced from YouTube and exhibits significant variations in camera motion, object appearance, and environmental conditions. As a result, UCF-101 ~\cite{soomro2012ucf101} serves as a robust benchmark for evaluating the performance of action recognition models.

\subsection{Dataset}
We evaluate our models on the UCF-101 dataset~\cite{soomro2012ucf101}.
%, a widely used benchmark for video action recognition containing 13,320 clips across 101 classes covering diverse daily and sports activities. The dataset exhibits large variations in motion, lighting, and background, making it a robust testbed for privacy-preserving federated training. 
We follow an 80–20 train–test split (9,537 training and 3,783 test videos). Each video is segmented into 16-frame clips, resized to $224\times224$ RGB using OpenCV~\cite{howse2013opencv} for compatibility with the MViT-B-16x4 architecture~\cite{fan2021multiscale}.

\begin{table}[t]
\centering
\caption{MViT-B-16x4 Architecture summary. Patch embedding uses Conv3d$(3,96,3\times7\times7)$ with positional encoding; selective tuning updates the final transformer stage and classification head.}
\label{tab:mvitb16x4}
\setlength{\tabcolsep}{3pt}
\begin{tabular}{lccc}
\hline
Stage & Dim & Blk & Stride (t,h,w) \\
\hline
Patch Emb. & 96  & 1 & (2,4,4) \\
Stage1    & 192 & 2 & (1,8,8)$\rightarrow$(1,4,4) \\
Stage2    & 384 & 3 & (1,4,4)$\rightarrow$(1,2,2) \\
Stage3    & 384 & 8 & (1,2,2) \\
Stage4    & 768 & 2 & (1,2,2)$\rightarrow$(1,1,1) \\
Norm  & 768 & - & - \\
Head  & 101 & - & - \\
\hline
\end{tabular}
\end{table}

% \begin{table*}[htbp]
% \centering
% \caption{Multiscale Vision Transformer (MViT-B-16x4) Architecture Details}
% \label{tab:mvitb16x4}
% \setlength{\tabcolsep}{2pt}
% \begin{tabular}{lccccl}
% \hline
% \textbf{Stage} & \textbf{Dim. In} & \textbf{Dim. Out} & \textbf{\# Blocks} & \textbf{Attention / Pooling Stride} & \textbf{Details} \\
% \hline
% Patch Embed & $3 \times 16 \times 224 \times 224$ & 96 & 1 & $(t, h, w)=(2, 4, 4)$ & Conv3d$(3,96,3{\times}7{\times}7)$ + PosEnc \\
% \hline
% Stage 1 & 96 & 192 & 2 & $(1, 8, 8) \rightarrow (1, 4, 4)$ & MSA with pool\_k,v stride 8, skip pool stride 2 \\
% Stage 2 & 192 & 384 & 3 & $(1, 4, 4) \rightarrow (1, 2, 2)$ & Dim doubling, skip pool stride 2 \\
% Stage 3 & 384 & 384 & 8 & $(1, 2, 2)$ & Core transformer body \\
% Stage 4 & 384 & 768 & 2 & $(1, 2, 2) \rightarrow (1, 1, 1)$ & Spatial resolution reduced, channel doubled \\
% \hline
% Normalization & 768 & 768 & - & - & LayerNorm$(768)$ \\
% Classification Head & 768 & 101 & - & - & Dropout(0.5) + Linear$(768,101)$ \\
% \hline
% \end{tabular}
% \end{table*}

\subsection{Experimental Framework}
We employ the pretrained \textbf{MViT-B-16x4}~\cite{fan2021multiscale} model (pretrained on Kinetics-400 and fine-tuned on UCF-101) to evaluate the privacy–utility–communication trade-offs of our proposed \textit{FedDP-STECAR}. 
Three experimental paradigms are considered:
(1) \textbf{Centralized DP training} to establish baselines with and without DP-SGD;
(2) \textbf{Federated DP training} using FedAvg and FedNova to study distributed performance; and 
(3) \textbf{Top-Level Sampling (TLS)}, which implements sampling without replacement for privacy amplification. 
Table~\ref{tab:mvitb16x4} summarizes the MViT-B-16x4 architecture used across all setups.

\textbf{Implementation Details}.
 We compare two fine-tuning strategies: full fine-tuning where all model parameters are updated during training, and selective fine-tuning where only specific layers (classification head and final transformer stages) are updated. The selective approach reduces computational overhead while improving privacy-utility trade-offs by limiting the parameter space.
%\textbf{Centralized Model DP Training.} 
%The baseline MViT-B-16x4 model~\cite{fan2021multiscale}, initially pretrained on the Kinetics-400 dataset, was fine-tuned on the UCF-101 dataset~\cite{soomro2012ucf101} using the \textit{Adam} optimizer (learning rate 0.001, batch size 8, 50 epochs).
%DP noise is injected via PyVacy~\cite{waites2019pyvacy}, selected for better MViT-B-16x4 layer compatibility than Opacus. Training uses micro-batch size 4 with $(\epsilon, \delta)$-DP where $\delta = 1e\text{-}4$ and $\epsilon = 0.65$ for strong privacy guarantees. The gradient clipping norm is set to $C = 1.0$, and video sampling remains deterministic. 
Both baseline and selective fine-tuning models with integrated TLS are trained under identical DP configurations for direct comparison.

\section{Results and Discussion}
% This section presents empirical evaluations of \textit{FedDP-STECAR} for privacy-preserving
% video activity recognition on the UCF101 dataset.
We analyze three key aspects aligned with our contributions:
(i) the privacy–utility trade-off of selective fine-tuning under differential privacy,
(ii) communication and computational efficiency in federated settings, and
(iii) aggregation-agnostic scalability across different federated optimizers.
\begin{figure}[h]
\centering
{\includegraphics[scale=0.5]{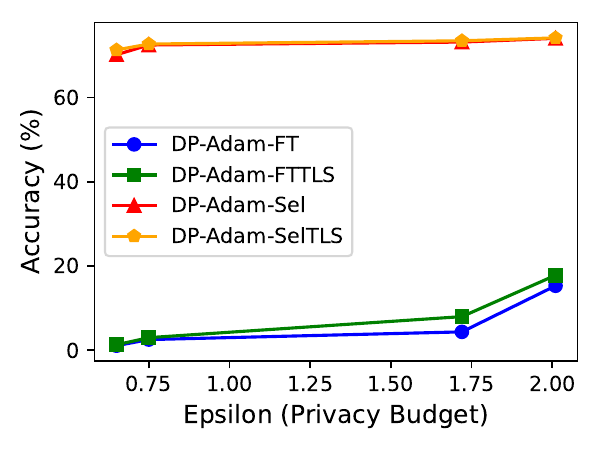}}

\caption{Centralized DP video activity recognition performance across varying $\epsilon$ (fixed $\delta = 1e\text{-}4$ and clipping norm $C=1.0$), using full fine-tuning, selective tuning, and integrated top-level sampling (without replacement). The baseline MViT-B-16x4 was pretrained on Kinetics-400 and fine-tuned on UCF101.}
\label{fig:centraldp}
\end{figure}
% =========================
% Table 1 — FedAvg: Comm & Compute Efficiency (Full vs Selective)
% =========================
\begin{table}[t]
\centering
% \caption{FedAvg on UCF101 at $\epsilon = 1.33$ ($\delta=10^{-4}$, $C=1.0$). Selective tuning (with/without TLS) slashes traffic and delay while improving accuracy and runtime. Two clients, 5 local epochs, 5 rounds; MViT-B-16x4 pretrained on Kinetics-400.}
\caption{FedAvg on UCF101 at $\epsilon = 1.33$ ($\delta=10^{-4}$, $C=1.0$); MViT-B-16x4 pretrained on Kinetics-400.}
\label{tab:fed_comm_eff}
\setlength{\tabcolsep}{4pt}
\resizebox{\linewidth}{!}{
\begin{tabular}{lcccc}
\toprule
\textbf{Model} & \textbf{Accuracy} (\%) & \textbf{Runtime} (min) & \textbf{Traffic} (MB) & \textbf{Comm. Delay} (s) \\
\midrule
DP-Adam-FT      & 22.42 & 339.38 & 1456 & 174.72 \\
DP-Adam-FTTLS   & 21.10 & 342.04 & 1456 & 174.72 \\
DP-Adam-Sel     & \textbf{73.06} & 179.68 & \textbf{3.10} & \textbf{0.37} \\
DP-Adam-SelTLS  & 72.58 & \textbf{179.64} & \textbf{3.10} & \textbf{0.37} \\
\bottomrule
\end{tabular}}
\end{table}

\subsection{Computation and Communication Efficiency}
\label{subsec:federated-efficiency}

We next examine \textit{FedDP-STECAR} in a federated setup under $\epsilon=1.33$,
using two clients, five local epochs, and five rounds of communication.
Table~\ref{tab:fed_comm_eff} compares full and selective fine-tuning models
under FedAvg. Selective tuning reduces runtime by nearly 47.5\%
while improving accuracy by more than 50 percentage points.
Importantly, communication traffic per client per round
drops from $1456$~MB to just $3.1$~MB---a reduction of $99.8$\%---and
communication latency decreases from $175$~s to $0.37$~s
(a $470\times$ speedup).
These results empirically confirm the communication cost model in
Eq.~\eqref{eq:joint_obj}, where the ratio $d_t/d \!\approx\! 0.0021$
aligns with the observed traffic compression.
The combination of selective tuning and TLS thus enables efficient
$\varepsilon$-regulation that balances privacy, utility, and bandwidth cost
for large-scale video models.

\subsection{Centralized DP Video Recognition}
\label{subsec:centralized}
To isolate the effects of selective tuning and TLS,
we first evaluate differentially private models in a centralized setup.
Selective tuning consistently improves performance across all privacy levels,
as shown in Fig.~\ref{fig:centraldp}.
At $\epsilon = 0.65$, \textit{DP-Adam-SelTLS} achieves
$70.21$\% higher accuracy than \textit{DP-Adam-FT} and $1.13$\% higher than
\textit{DP-Adam-Sel}, confirming that confining DP noise to fine-tuned layers
preserves utility under tight privacy budgets.
These results validate the effectiveness of
\textit{FedDP-STECAR}'s selective fine-tuning and TLS strategies in achieving better
privacy–utility trade-offs compared to full fine-tuning.

% =========================
% Table 2 — Aggregation-Agnostic Behavior (2 clients, ε=1.33)
% =========================
\begin{table}[h]
\centering
% \caption{Aggregation-agnostic evaluation at $\epsilon = 1.33$ ($\delta=10^{-4}$) with two clients and 10 rounds. Selective tuning achieves $\sim$2$\times$ faster training and $>45$ pp accuracy gains over full fine-tuning; FedAvg and FedNova are comparable.}
\caption{FedAvg vs. FedNova at $\epsilon{=}1.33$ ($\delta{=}10^{-4}$): selective tuning improves accuracy and halves runtime.}

\label{tab:agg_agnostic_2c}
\setlength{\tabcolsep}{6pt}
\begin{tabular}{lccc}
\toprule
\textbf{Model} & $\boldsymbol{\epsilon}$ & \textbf{Accuracy} (\%) & \textbf{Runtime} (min) \\
\midrule
DP-Adam-FTFA   & 1.33 & 28.3 & 681.16 \\
DP-Adam-FTFN   & 1.33 & 28.7 & 684.93 \\
DP-Adam-SelFA  & 1.33 & \textbf{74.1} & \textbf{359.54} \\
DP-Adam-SelFN  & 1.33 & 73.5 & 360.31 \\
\bottomrule
\end{tabular}
\end{table}

\subsection{Aggregation-Agnostic Federated Learning}
\label{subsec:aggregation-agnostic}

To evaluate the generality of \textit{FedDP-STECAR}, we compare its performance across
two popular federated optimizers: \textit{FedAvg} and \textit{FedNova}.
As shown in Table~\ref{tab:agg_agnostic_2c}, selective fine-tuning achieves
nearly identical accuracy and runtime across both aggregation schemes.
At $\epsilon=1.33$ with two clients and ten rounds,
\begin{table}[h]
\centering
% \caption{Scalability across clients/rounds and privacy budgets ($\delta=10^{-4}$, $C=1.0$). Selective tuning remains stable as clients increase; full fine-tuning degrades significantly. Traffic per round remains $\approx$3.1 MB with selective tuning (omitted to save space).}
\caption{Scalability across clients, rounds, and privacy budgets ($\delta{=}10^{-4}$, $C{=}1.0$).}
\label{tab:scalability}
\setlength{\tabcolsep}{3pt}
\resizebox{\linewidth}{!}{
\begin{tabular}{ccccccc}
\toprule
\textbf{Clients} & \textbf{Rounds} & $\boldsymbol{\epsilon}$ & \textbf{FT-FA} (\%) & \textbf{Sel-FA} (\%) & \textbf{FT-FN} (\%) & \textbf{Sel-FN} (\%) \\
\midrule
2  & 10 & 1.33 & 28.3 & \textbf{74.1} & 28.7 & 73.5 \\
\midrule
5  & 20 & 1.33 & 30.6 & 73.8 & 32.0 & \textbf{74.0} \\
10 & 20 & 1.42 & 27.5 & \textbf{73.7} & 27.9 & 73.5 \\
20 & 20 & 3.15 & 18.3 & \textbf{72.5} & 19.1 & \textbf{72.5} \\
\bottomrule
\end{tabular}}
\end{table}

\textit{DP-Adam-SelFA} reaches $74.1$\% accuracy versus $73.5$\% for
\textit{DP-Adam-SelFN}, demonstrating that our framework is indeed
aggregation-agnostic and compatible with various federated strategies.
Full fine-tuning baselines remain significantly lower ($<30$\% accuracy),
highlighting the sensitivity of high-dimensional gradient updates under DP noise.

% =========================
% Table 3 — Scalability: Clients, Rounds, and Privacy Budgets
% =========================

\subsection{Scalability with Client and Privacy Variations}
\label{subsec:scalability}

Finally, Table~\ref{tab:scalability} evaluates scalability across client counts
($2\!\to\!20$), communication rounds ($10\!\to\!20$), and privacy budgets
($\epsilon\!\in\!\{1.33,1.42,3.15\}$).
Selective tuning maintains accuracy above $72$\% even at 20 clients and
$\epsilon=3.15$, whereas full fine-tuning accuracy collapses to below $20$\%.
The modest 1–2\% degradation across larger client pools demonstrates that
\textit{FedDP-STECAR} scales effectively under stricter communication and privacy constraints.
Additionally, the communication cost remains nearly constant ($\sim3$~MB per round).

\section{Conclusion}

This work presents \textbf{FedDP-STECAR}, a framework that jointly addresses privacy, communication efficiency and accuracy in federated video-based human action recognition. By combining selective fine-tuning with differential privacy, we mitigate model exposure through targeted perturbations of high-impact layers while preserving temporal coherence and reducing communication overhead. Our aggregation-agnostic design generalizes across federated protocols (FedAvg, FedNova) and maintains high accuracy under strict privacy budgets  ($\epsilon\leq 1.33$), with Top-Level Sampling providing privacy amplification in centralized scenarios.

%\section{REFERENCES}
%\label{sec:refs}

\bibliographystyle{IEEEbib}
\bibliography{refoo,refs,strings}

\end{document}